\documentclass[10pt,twocolumn,letterpaper]{article}

\usepackage{cvpr}      

\usepackage[dvipsnames]{xcolor}

\usepackage{textcomp}
\definecolor{cvprblue}{rgb}{0.21,0.49,0.74}
\usepackage[pagebackref,breaklinks,colorlinks,citecolor=cvprblue]{hyperref}
\usepackage{algorithm}
\usepackage{algpseudocode}
\usepackage{xcolor}
\definecolor{lightgray}{gray}{0.5} 

\def\paperID{12943} 
\def\confName{CVPR}
\def\confYear{2024}

\title{TI2V-Zero: Zero-Shot Image Conditioning for Text-to-Video Diffusion Models
}

\author{
Haomiao Ni$^1$\thanks{Work done during an internship at MERL.}
\qquad Bernhard Egger$^2$
\qquad Suhas Lohit$^3$
\qquad Anoop Cherian$^3$
\qquad Ye Wang$^3$\\
Toshiaki Koike-Akino$^3$
\qquad Sharon X. Huang$^1$
\qquad Tim K. Marks$^3$ \\
\small{$^1$The Pennsylvania State University, USA 
\qquad$^2$Friedrich-Alexander-Universit\"{a}t Erlangen-N\"{u}rnberg}, Germany\\
\small{$^3$Mitsubishi Electric Research Laboratories (MERL), USA
}\\
\scriptsize{
$^1${\tt\{hfn5052, suh972\}@psu.edu}
\quad$^2${\tt bernhard.egger@fau.de}
\quad$^3${\tt\{slohit,cherian,yewang,koike,tmarks\}@merl.com}
}\\
\small{\url{https://merl.com/demos/TI2V-Zero}}
}

\usepackage{pifont}
\newcommand{\cmark}{\ding{51}}%
\newcommand{\xmark}{\ding{55}}%
\usepackage{multirow}

\newlength{\myabovedisplayskip}
\newlength{\mybelowdisplayskip}
\BeforeBeginEnvironment{equation}{\setlength{\abovedisplayskip}{\myabovedisplayskip}\setlength{\belowdisplayskip}{\mybelowdisplayskip}}
\AfterEndEnvironment{equation}{\setlength{\abovedisplayskip}{\myabovedisplayskip}\setlength{\belowdisplayskip}{\mybelowdisplayskip}}

\begin{document}
\maketitle
\begin{abstract}
Text-conditioned image-to-video generation (TI2V) aims to synthesize a realistic video starting from a given image (\eg, a woman's photo) and a text description (\eg, ``a woman is drinking water."). Existing TI2V frameworks often require costly training on video-text datasets and specific model designs for text and image conditioning. In this paper, we propose TI2V-Zero, a zero-shot, tuning-free method that empowers a pretrained text-to-video (T2V) diffusion model to be conditioned on a provided image, enabling TI2V generation without any optimization, fine-tuning, or introducing external modules. Our approach leverages a pretrained T2V diffusion foundation model as the generative prior. To guide video generation with the additional image input, we propose a ``repeat-and-slide'' strategy that modulates the reverse denoising process, allowing the frozen diffusion model to synthesize a video frame-by-frame starting from the provided image. To ensure temporal continuity, we employ a DDPM inversion strategy to initialize Gaussian noise for each newly synthesized frame and a resampling technique to help preserve visual details. We conduct comprehensive experiments on both domain-specific and open-domain datasets, where TI2V-Zero consistently outperforms a recent open-domain TI2V model. Furthermore, we show that TI2V-Zero can seamlessly extend to other tasks such as video infilling and prediction when provided with more images. Its autoregressive design also supports long video generation.
\end{abstract}

\vspace{-1em}
\section{Introduction}
\noindent
Image-to-video (I2V) generation is an appealing topic with various applications, including artistic creation, entertainment, and data augmentation for machine learning~\cite{ni2023conditional}. 
Given a single image $x^0$ and a text prompt $y$, text-conditioned image-to-video (TI2V) generation aims to synthesize $M$ new frames to yield a realistic video, $\hat{\bf{x}}=\langle x^0, \hat{x}^1, \dots, \hat{x}^M \rangle$, starting from the given frame $x^0$ and satisfying the text description $y$. 
Current TI2V generation methods~\cite{wang2023videocomposer,yin2023dragnuwa,villegas2022phenaki} typically rely on computationally-heavy training on video-text datasets and specific architecture designs to enable text and image conditioning.
Some~\cite{fu2023tell,hu2022make} are constrained to specific domains due to the lack of training with large-scale open-domain datasets. Other approaches, such as~\cite{xing2023dynamicrafter,guo2023animatediff}, utilize pretrained foundation models to reduce training costs, but they still need to train additional modules using video data.

\begin{figure}[t]
    \centering
    \includegraphics[width=0.9\linewidth]{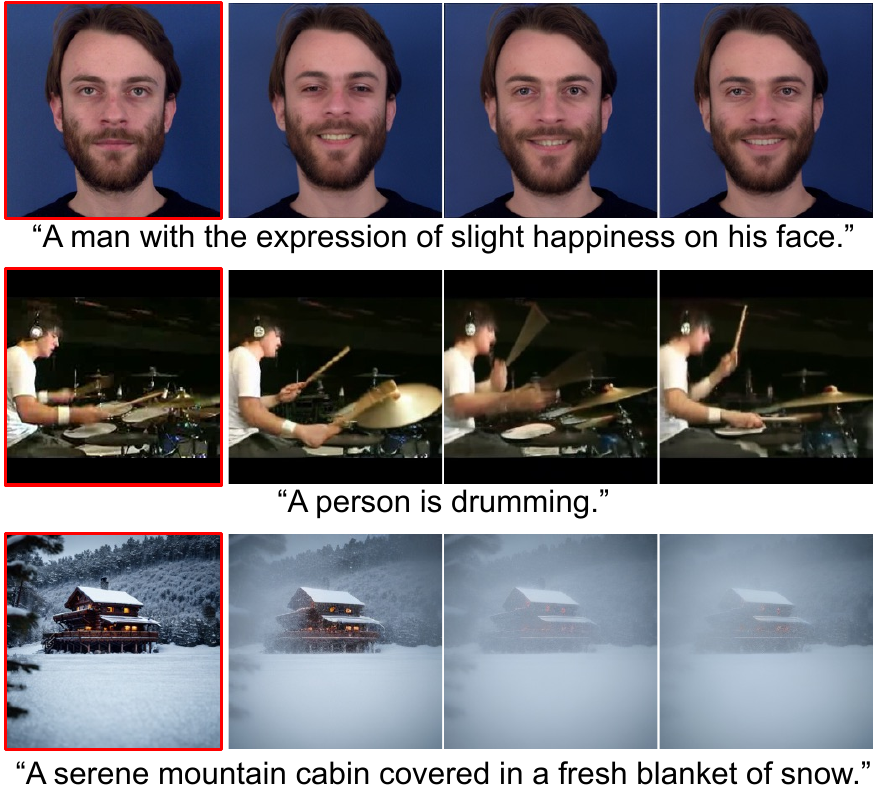}
    \caption{Examples of generated video frames using our proposed TI2V-Zero.
    The given first image $x^0$ is highlighted with the red box, and the text condition $y$ is shown under each row of the video. The remaining columns show the 6th, 11th, and 16th frames of the generated output videos. Each generated video has 16 frames with a resolution of $256\times256$.
    }
    \label{fig:showcase}
    \vspace{-1em}
\end{figure}

In this paper, we propose TI2V-Zero, which achieves \textit{zero-shot} TI2V generation using only an open-domain pretrained text-to-video (T2V) latent diffusion model \cite{wang2023modelscope}.
Here ``zero-shot'' means that when using the diffusion model (DM) that was trained only for text conditioning, our framework enables image conditioning without any optimization, fine-tuning, or introduction of additional modules.
Specifically, we guide the generation process by incorporating the provided image $x^0$ into the output latent code at each reverse denoising step. 
To ensure that the temporal attention layers of the pretrained DM focus on information from the given image, we propose a ``repeat-and-slide'' strategy to synthesize the video in a frame-by-frame manner, rather than directly generating the entire video volume. 
Notably, TI2V-Zero is not trained for the specific domain of the provided image, thus allowing the model to generalize to any image during inference. Additionally, its autoregressive generation makes the synthesis of long videos possible. 

While the standard denoising sampling process starting with randomly initialized Gaussian noise can produce matching semantics, it often results in temporally inconsistent videos. 
Therefore, we introduce an inversion strategy based on the DDPM \cite{ho2020denoising} forward process, to provide a more suitable initial noise for generating each new frame. 
We also apply a resampling technique \cite{lugmayr2022repaint} in the video DM to help preserve the generated visual details. Our approach ensures that the network maintains temporal consistency, generating visually convincing videos conditioned on the given starting image (see \cref{fig:showcase}). 

We conduct extensive experiments on MUG \cite{aifanti2010mug}, UCF-101 \cite{soomro2012ucf101}, and a new open-domain dataset. In these experiments, TI2V-Zero consistently performs well, outperforming a state-of-the-art model \cite{xing2023dynamicrafter} that was based on a video diffusion foundation model \cite{chen2023videocrafter1} and was specifically trained to enable open-domain TI2V generation.

\section{Related Work}
\subsection{Conditional Image-to-Video Generation}
\noindent
Conditional video generation aims to synthesize videos guided by user-provided signals.
It can be classified according to which type(s) of conditions are given, such as text-to-video (T2V) generation \cite{hong2022cogvideo,li2018video,wu2021godiva,blattmann2023align,he2022latent,ho2022imagen}, video-to-video (V2V) generation \cite{ni20243d,chan2019everybody,ni2023cross,wang2018video,qi2023fatezero,wang2022latent}, and image-to-video (I2V) generation \cite{blattmann2021understanding,dorkenwald2021stochastic,hu2022make,mahapatra2022controllable,ni2023conditional,yang2018pose}. 
Here we discuss previous text-conditioned image-to-video (TI2V) generation methods \cite{fu2023tell,ho2022video,guo2023animatediff,wang2023videocomposer,yin2023dragnuwa,preechakul2022diffusion}. Hu \etal \cite{hu2022make} introduced MAGE, a TI2V generator that integrates a motion anchor structure to store appearance-motion-aligned representations through three-dimensional axial transformers. 
Yin \etal \cite{yin2023dragnuwa} proposed DragNUWA, a diffusion-based model capable of generating videos controlled by text, image, and trajectory information with three modules including a trajectory sampler, a multi-scale fusion, and an adaptive training strategy.
However, these TI2V frameworks require computationally expensive training on video-text datasets and a particular model design to support text-and-image-conditioned training. 
In contrast, our proposed TI2V-Zero leverages a pretrained T2V diffusion model to achieve zero-shot TI2V generation without additional optimization or fine-tuning, making it suitable for a wide range of applications.  

\subsection{Adaptation of Diffusion Foundation Models}
Due to the recent successful application of diffusion models (DM) \cite{ ho2020denoising, nichol2021improved, rombach2022high, sohl2015deep, song2019generative} to both image and video generation, visual diffusion foundation models have gained prominence. 
These include text-to-image (T2I) models such as Imagen \cite{saharia2022photorealistic} and Stable Diffusion \cite{rombach2022high}, as well as text-to-video (T2V) models such as ModelScopeT2V \cite{wang2023modelscope} and VideoCrafter1 \cite{chen2023videocrafter1}.
These models are trained with large-scale open-domain datasets, often including LAION-400M \cite{schuhmann2021laion} and WebVid-10M \cite{Bain21}. 
They have shown immense potential for adapting their acquired knowledge base to address a wide range of downstream tasks, thereby reducing or eliminating the need for extensive labeled data.
For example, previous works have explored the application of large T2I models to personalized image generation \cite{gal2022image,ruiz2022dreambooth}, image editing \cite{hertz2022prompt,mokady2022null,lugmayr2022repaint,meng2021sdedit,nair2023steered}, image segmentation \cite{xu2023open,baranchuk2021label}, video editing \cite{vid2vid-zero, qi2023fatezero}, and video generation \cite{wu2023tune, khachatryan2023text2video,singer2022make, guo2023animatediff}. 
In contrast to T2I models, there are fewer works on the adaptation of large-scale T2V models. 
Xing \etal \cite{xing2023dynamicrafter} proposed DynamicCrafter for open-domain TI2V generation by adapting a T2V foundation model \cite{chen2023videocrafter1}. To control the generative process, they first employed a learnable image encoding network to project the given image into a text-aligned image embedding space. Subsequently, they utilized dual cross-attention layers to fuse text and image information and also concatenated the image with the initial noise to provide the video DM with more precise image details. 
In contrast, in this paper we explore how to inject the provided image to guide the DM sampling process based solely on the pretrained T2V model itself, with no additional training for the new TI2V task. 

\begin{figure*}[t]
    \centering
\includegraphics[width=\textwidth]{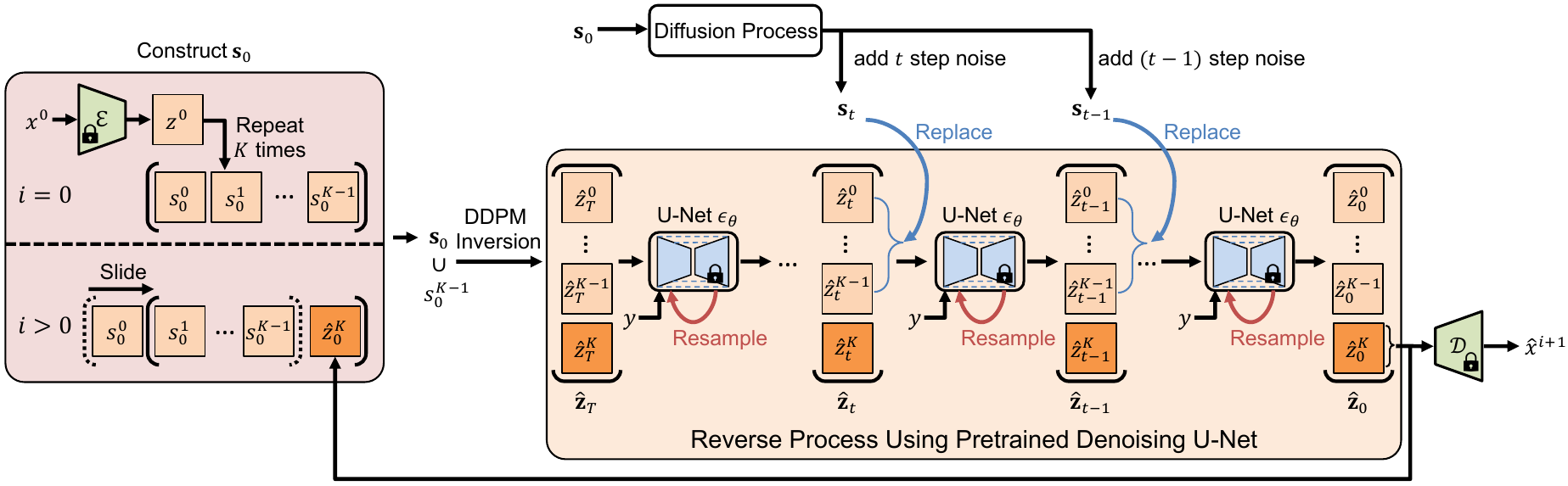}
    \caption{Illustration of the process of applying TI2V-Zero to generate the new frame $\hat{x}^{i+1}$, given the starting image $x^0$ and text $y$. TI2V-Zero is built upon a \textit{frozen} pretrained T2V diffusion model, including frame encoder $\mathcal{E}$, frame decoder $\mathcal{D}$, and the denoising U-Net $\epsilon_\theta$. At the beginning of generation ($i=0$), we encode $x^0$ as $z^0$ and repeat it $K$ times to form the queue $\mathbf{s}_0$. We then apply DDPM-based inversion to $\mathbf{s}_0$ to produce the initial Gaussian noise $\hat{\mathbf{z}}_T$. Subsequently, in each reverse denoising step using U-Net $\epsilon_\theta$, we keep replacing the first $K$ frames of $\hat{\mathbf{z}}_t$ with the noisy latent code $\mathbf{s}_t$ derived from $\mathbf{s}_0$. 
    Resampling is also applied within each step to improve motion coherence.
    We finally decode the final frame of the clean latent code $\hat{\mathbf{z}}_0$ as the new synthesized frame $\hat{x}^{i+1}$. To compute the new $\mathbf{s}_0$ for the next iteration of generation ($i>0$), we perform a {sliding} operation by dequeuing ${s}^0_0$ and enqueuing $\hat{z}^K_0$ within $\mathbf{s}_0$. 
    }
    \label{fig:framework}
\end{figure*}

\section{Methodology}
Given one starting image $x^0$ and text $y$, let $\mathbf{x}=\langle x^0, x^1, \dots, x^M\rangle$ represent a real video corresponding to text $y$.
The objective of text-conditioned image-to-video (TI2V) generation is to synthesize a video $\hat{\mathbf{x}}=\langle{x}^0, \hat{x}^1, \dots, \hat{x}^M\rangle$, such that the conditional distribution of $\hat{\mathbf{x}}$ given $x^0$ and $y$ is identical to the conditional distribution of $\mathbf{x}$ given $x^0$ and $y$, i.e., $p(\hat{\mathbf{x}}|x^0, y)=p(\mathbf{x}|x^0, y)$. 
Our proposed TI2V-Zero can be built on a pretrained T2V diffusion model with a 3D-UNet-based denoising network. Here we choose ModelScopeT2V \cite{wang2023modelscope} as backbone due to its promising open-domain T2V generation ability.
Below, we first introduce preliminaries about diffusion models, then introduce the architecture of the pretrained T2V model, and finally present the details of our TI2V-Zero.

\subsection{Preliminaries: Diffusion Models}
Diffusion Models (DM) \cite{ho2020denoising,sohl2015deep,song2019generative} are probabilistic models designed to learn a data distribution. 
Here we introduce the fundamental concepts of Denoising Diffusion Probabilistic Models (DDPM).
Given a sample from the data distribution $\mathbf{z}_0\sim q(\mathbf{z}_0)$, the \textit{forward} diffusion process of a DM produces a Markov chain $\mathbf{z}_1, \dots, \mathbf{z}_T$ by iteratively adding Gaussian noise to $\mathbf{z}_0$ according to a variance schedule $\beta_1, \dots, \beta_T$, that is:
\begin{small}
\begin{equation}
\label{eq:forward}
    q(\mathbf{z}_t|\mathbf{z}_{t-1}) = \mathcal{N}(\mathbf{z}_t; \sqrt{1-\beta_t}\mathbf{z}_{t-1}, \beta_t\mathbf{I})
\enspace,
\end{equation}
\end{small}
where variances $\beta_t$ are constant.
When the $\beta_t$ are small, the posterior $q(\mathbf{z}_{t-1}|\mathbf{z}_{t})$ can be well approximated by a diagonal Gaussian \cite{sohl2015deep,nichol2021glide}. 
Furthermore, if the length of the chain, denoted by $T$, is sufficiently large, $\mathbf{z}_T$ can be well approximated by a standard Gaussian distribution $\mathcal{N}(\mathbf{0}, \mathbf{I})$. 
These suggest that the true posterior $q(\mathbf{z}_{t-1}|\mathbf{z}_{t})$ can be estimated by $p_\theta(\mathbf{z}_{t-1}|\mathbf{z}_t)$ defined as:
\begin{small}
\begin{equation}
\label{eq:reverse}
    p_\theta(\mathbf{z}_{t-1}|\mathbf{z}_t)=\mathcal{N}(\mathbf{z}_{t-1}; \mu_\theta(\mathbf{z}_t), \sigma_t^2\mathbf{I})
\enspace,
\end{equation}
\end{small}
where variances $\sigma_t$ are also constants.
The \textit{reverse} denoising process in the DM (also termed \textit{sampling}) then generates samples $\mathbf{z}_0\sim p_\theta(\mathbf{z}_0)$ by starting with Gaussian noise $\mathbf{z}_T\sim \mathcal{N}(\mathbf{0}, \mathbf{I})$ and gradually reducing noise in a Markov chain $\mathbf{z}_{T-1}, \mathbf{z}_{T-2}, \dots, \mathbf{z}_0$ using a learned $p_\theta(\mathbf{z}_{t-1}|\mathbf{z}_t)$. 
To learn $p_\theta(\mathbf{z}_{t-1}|\mathbf{z}_t)$, Gaussian noise $\epsilon$ is first added to $\mathbf{z}_0$ to generate samples $\mathbf{z}_t$. 
Utilizing the independence property of the noise added at each forward step in \cref{eq:forward}, we can calculate the total noise variance as $\Bar{\alpha}_t=\prod^t_{i=0}(1-\beta_i)$ and transform $\mathbf{z}_0$ to $\mathbf{z}_t$ in a single step:
\begin{small}
\begin{equation}
    \label{eq:single_forward}
    q(\mathbf{z}_t|\mathbf{z}_0)= \mathcal{N}(\mathbf{z}_t; \sqrt{\Bar{\alpha}_t}\mathbf{z}_0, (1-\Bar{\alpha}_t)\mathbf{I})
    \enspace.
\end{equation}
\end{small}
Then a model $\epsilon_\theta$ is trained to predict $\epsilon$ using the following mean-squared error loss:
\begin{small}
\begin{equation}
    L=\mathbb{E}_{t\sim \mathcal{U}(1, T), \mathbf{z}_0\sim q(\mathbf{z}_0), \epsilon\sim \mathcal{N}(\mathbf{0}, \mathbf{I})}\left[||\epsilon-\epsilon_\theta(\mathbf{z}_t, t)||^2\right]
\enspace,
\end{equation}
\end{small}
where diffusion step $t$ is uniformly sampled from $\{1, \dots, T\}$. 
Then $\mu_\theta(\mathbf{z}_t)$ in \cref{eq:reverse} can be derived from $\epsilon_\theta(\mathbf{z}_t, t)$ to model $p_\theta(\mathbf{z}_{t-1}|\mathbf{z}_t)$ \cite{ho2020denoising}.
The denoising model $\epsilon_\theta$ is implemented using a time-conditioned U-Net \cite{ronneberger2015u} with residual blocks \cite{he2016deep} and self-attention layers \cite{vaswani2017attention}. Diffusion step $t$ is specified to $\epsilon_\theta$ by the sinusoidal position embedding \cite{vaswani2017attention}.
Conditional generation that samples \mbox{$\mathbf{z}_0 \sim p_\theta(\mathbf{z}_0|y)$} can be achieved by learning a $y$-conditioned model $\epsilon_\theta(\mathbf{z}_t, t, y)$ \cite{nichol2021glide,rombach2022high} with \textit{classifier-free} guidance~\cite{ho2021classifier}.
During training, the condition $y$ in $\epsilon_\theta(\mathbf{z}_t, t, y)$ is replaced by a null label $\emptyset$ with a fixed probability. When sampling, the output is generated as follows:
\vspace{-1em}
\begin{small}
\begin{equation}
\label{eq:classifier}
    \hat{\epsilon}_\theta(\mathbf{z}_t, t, y) = \epsilon_\theta(\mathbf{z}_t, t, \emptyset) + g \cdot (\epsilon_\theta(\mathbf{z}_t, t, y)-\epsilon_\theta(\mathbf{z}_t, t, \emptyset))
\enspace,
\end{equation}
\end{small}
where $g$ is the guidance scale.

\begin{algorithm}[t]
\small
    \caption{Generation using our TI2V-Zero approach.
    }
    \begin{algorithmic}[1]
    \algrenewcommand\algorithmiccomment[1]{\hfill\textcolor{lightgray}{// #1}} 
    \algnewcommand{\LineComment}[1]{\Statex \textcolor{lightgray}{// #1}}
    \Require The starting frame $x^0$; The text prompt $y$; The pretrained T2V Model $\mathcal{M}$ for generating $(K+1)$-frame videos, including frame encoder $\mathcal{E}$ and frame decoder $\mathcal{D}$, and the DM denoising networks $\epsilon_\theta$; 
    The iteration number $U$ for resampling;
    The parameter $M$ to control the length of the output video.
    \Ensure A synthesized video $\hat{\mathbf{x}}$ with $(M+1)$ frames.
    \State $z^0\gets \mathcal{E}(x^0)$
    \Comment{Encode $x^0$}
    \State $\mathbf{s}_0 \gets \langle z^0, z^0, \cdots, z^0\rangle$
    \Comment{Repeat $z^0$ for $K$ times}
    \State $\hat{\mathbf{x}}\gets\langle x^0\rangle$
    \For{$i=1, 2, \cdots, M$}
        \Statex \textcolor{lightgray}{\hspace*{\algorithmicindent}// Generate one new frame $\hat{x}^i$}
        \State $\mathbf{s}_T\sim\mathcal{N}(\sqrt{\Bar{\alpha}}_{T}\mathbf{s}_0, (1-\Bar{\alpha}_{T})\mathbf{I})$ \Comment{DDPM Inversion}
        \State $\hat{z}^K_T\sim\mathcal{N}(\sqrt{\Bar{\alpha}}_{T}{s}^{K-1}_0, (1-\Bar{\alpha}_{T})\mathbf{I})$
        \State $\hat{\mathbf{z}}_T\gets \mathbf{s}_T\cup \hat{z}^K_T$\Comment{Initialize $\hat{\mathbf{z}}_T$}
        \For{$t=T-1, \cdots, 2, 1$}
            \State $\mathbf{s}_{t}\sim\mathcal{N}(\sqrt{\Bar{\alpha}}_{t}\mathbf{s}_0, (1-\Bar{\alpha}_{t})\mathbf{I})$
            \For{$u=1, 2, \cdots, U$}
            \State $\langle\hat{z}_{t}^0, \hat{z}_{t}^1, \cdots, \hat{z}_{t}^{K-1}\rangle\gets \mathbf{s}_{t}$\Comment{Replace}
            \State $\hat{\mathbf{z}}_{t-1}\sim\mathcal{N}(\mu_\theta(\hat{\mathbf{z}}_t, y), \sigma_t^2\mathbf{I})$
            \If{$u<U$ and $t>1$}
            \State $\hat{\mathbf{z}}_t\sim\mathcal{N}(\sqrt{1-\beta_t}\hat{\mathbf{z}}_{t-1}, \beta_t\mathbf{I})$\Comment{Resample} 
            \EndIf
            \EndFor
        \EndFor
        \State $\mathbf{s}_0\gets\langle s^1_0, s^2_0, \cdots, s^{K-1}_0\rangle\cup\hat{z}^K_0$ \Comment{Slide}
        \State $\hat{x}^i\gets\mathcal{D}(\hat{z}^K_0)$ \Comment{Decode $\hat{z}^K_0$}
        \State $\hat{\mathbf{x}}\gets\hat{\mathbf{x}}\cup\hat{x}^i$ 
    \EndFor
    \State \Return $\hat{\mathbf{x}}$
    \end{algorithmic}
\label{alg:repeating_and_sliding}
\end{algorithm}

\subsection{Architecture of Pretrained T2V Model}
TI2V-Zero can be built upon a pretrained T2V diffusion model with a 3D-UNet-based denoising network. Here we choose ModelScopeT2V \cite{wang2023modelscope} as the pretrained model (denoted $\mathcal{M}$). 
We now describe this T2V model in detail.

\textbf{Structure Overview.}  Given a text prompt $y$, the T2V model $\mathcal{M}$ synthesizes a video $\hat{\mathbf{x}}=\langle\hat{x}^0, \hat{x}^1, \dots, \hat{x}^K\rangle$ with a pre-defined video of length $(K+1)$ using a latent video diffusion model. 
Similar to Latent Diffusion Models (LDM) \cite{rombach2022high}, $\mathcal{M}$ incorporates a frame auto-encoder \cite{kingma2013auto,esser2021taming} for the conversion of data between pixel space $\mathcal{X}$ and latent space $\mathcal{Z}$ through its encoder $\mathcal{E}$ and decoder $\mathcal{D}$. 
Given the real video $\mathbf{x}=\langle x^0, x^1, \dots, x^K\rangle$, $\mathcal{M}$ first utilizes the frame encoder $\mathcal{E}$ to encode the video $\mathbf{x}$ as $\mathbf{z}=\langle z^0, z^1, \dots, z^K\rangle$. 
Here the sizes of pixel frame $x$ and latent frame $z$ are $H_x\times W_x\times 3$ and $H_z\times W_z\times C_z$, respectively. 
To be consistent with the notation used for the DM, we denote the clean video latent $\mathbf{z}=\mathbf{z}_0=\langle z_0^0, z_0^1, \dots, z_0^K\rangle$.
$\mathcal{M}$ then learns a DM on the latent space $\mathcal{Z}$ through a 3D denoising U-Net $\epsilon_\theta$ \cite{cciccek20163d}. 
Let $\mathbf{z}_t=\langle {z}_t^0, {z}_t^1, \dots, {z}_t^K\rangle$ represent the latent sequence that results from adding noise over $t$ steps to the original latent sequence $\mathbf{z}_0$.
When training, the forward diffusion process of a DM transforms the initial latent sequence $\mathbf{z}_0$ into $\mathbf{z}_T$ by iteratively adding Gaussian noise $\epsilon$ for $T$ steps. 
During inference, denoising U-Net $\epsilon_\theta$ predicts the added noise at each step, enabling the generation of the clean latent sequence $\hat{\mathbf{z}}_0=\langle \hat{z}_0^0, \hat{z}_0^1, \dots, \hat{z}_0^K\rangle$ starting from randomly sampled Gaussian noise $\mathbf{z}_T\sim \mathcal{N}(\mathbf{0}, \mathbf{I})$.

\textbf{Text Conditioning Mechanism.} 
$\mathcal{M}$ employs a cross-attention mechanism \cite{rombach2022high} to incorporate text information into the generative process as guidance.
Specifically, $\mathcal{M}$ uses a pretrained CLIP model~\cite{radford2021learning} to encode the prompt $y$ as the text embedding $e$. The embedding $e$ is later used as the key and value in the multi-head attention layer within the spatial attention blocks, thus enabling the integration of text features with the intermediate U-Net features in $\epsilon_\theta$.

\textbf{Denoising U-Net.} 
The denoising U-Net $\epsilon_\theta$ includes four key building blocks: the initial block, the downsampling block, the spatio-temporal block, and the upsampling block.
The initial block transfers the input into the embedding space, while the downsampling and upsampling blocks are responsible for spatially downsampling and upsampling the feature maps. 
The spatio-temporal block is designed to capture spatial and temporal dependencies in the latent space, which comprises 2D spatial convolution, 1D temporal convolution, 2D spatial attention, and 1D temporal attention.

\begin{figure}[t]
    \centering
    \includegraphics[width=0.92\linewidth]{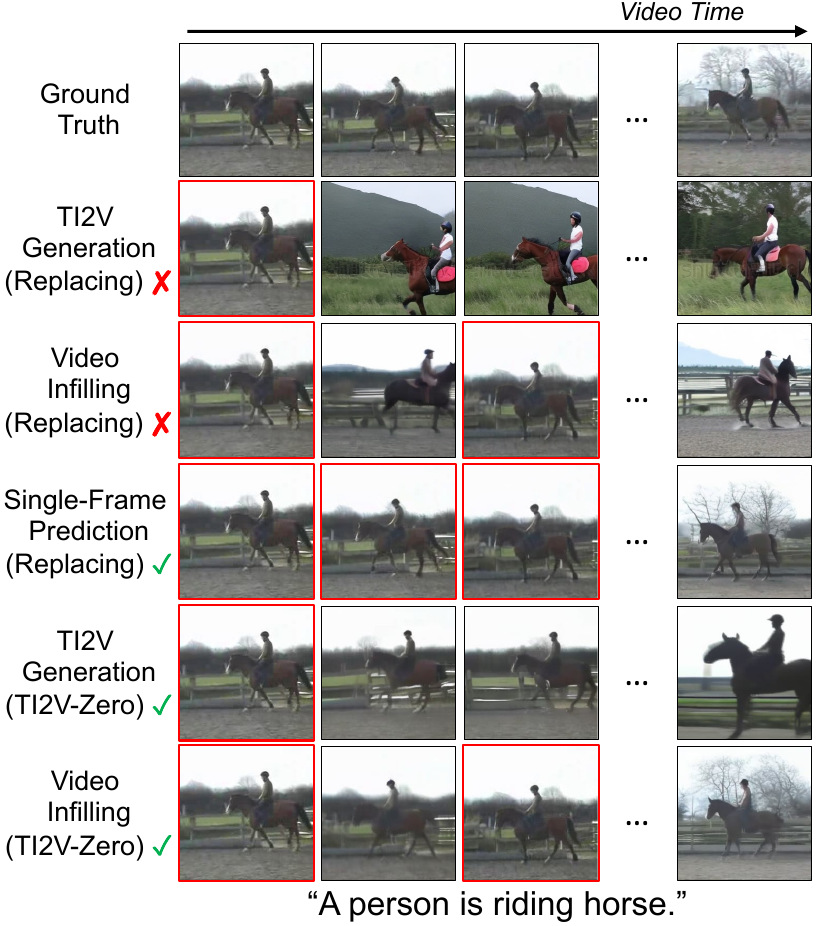}
    \caption{
    Illustration of the motivation behind our framework. We explore the application of a replacing-based baseline approach (rows 2--4, labeled ``Replacing'') and our TI2V-Zero (rows 5--6, labeled ``TI2V-Zero'') in various video generation tasks.
    The given real frames for each task are highlighted by red boxes and the text input is shown under the block.
    The replacing-based approach is only effective at predicting a single frame when all the other frames in the video are provided, while TI2V-Zero generates temporally coherent videos for both the TI2V and video infilling tasks.
    }
    \label{fig:motivation}
    \vspace{-1em}
\end{figure}

\subsection{Our Framework}
Leveraging the pretrained T2V foundation model $\mathcal{M}$, we first propose a straightforward \textit{replacing}-based baseline for adapting $\mathcal{M}$ to TI2V generation. We then analyze the possible reasons why it fails and introduce our TI2V-Zero framework, which includes a repeat-and-slide strategy, DDPM-based inversion, and resampling. Figure \ref{fig:framework} and \cref{alg:repeating_and_sliding} demonstrate the inference process of TI2V-Zero.

\textbf{Replacing-based Baseline.}
We assume that the pretrained model $\mathcal{M}$ is designed to generate the video with a fixed length of $(K+1)$. So we first consider synthesizing videos with that same length $(K+1)$, i.e., $M=K$.
Since the DM process operates within the latent space $\mathcal{Z}$, we use the encoder $\mathcal{E}$ to map the given starting frame $x^0$ into the latent representation $z^0$. 
Additionally, we denote $z^0=z^0_0$ to specify that the latent is clean and corresponds to diffusion step~0 of the DM.
Note that each reverse denoising step in \cref{eq:reverse} from $\hat{\mathbf{z}}_t$ to $\hat{\mathbf{z}}_{t-1}$ depends solely on $\hat{\mathbf{z}}_t=\langle\hat{z}_t^0, \hat{z}_t^1, \dots, \hat{z}_t^K\rangle$.
To ensure that the first frame of the final synthesized clean video latent $\hat{\mathbf{z}}_0=\langle \hat{z}_0^0, \hat{z}_0^1, \dots, \hat{z}_0^K\rangle$ at step 0 matches the provided image latent, i.e., $\hat{z}_0^0=z^0_0$,
we can modify the first generated latent $\hat{z}^0_t$ of $\hat{\mathbf{z}}_t$ at each reverse step, as long as the signal-to-noise ratio of each frame latent in $\hat{\mathbf{z}}_t$ remains consistent.
Using \cref{eq:single_forward}, we can add $t$ steps of noise to the provided image latent $z^0_0$, allowing us to sample $z^0_t$ through a single-step calculation.
By \textit{replacing} the first generated latent $\hat{z}^0_t$ with the noisy image latent $z^0_t$ at each reverse denoising step, we might expect that the video generation process can be guided by $z^0_0$ with the following expressions defined for each reverse step:
\begin{small}
\begin{subequations}
\label{eq:naive_sampling}
    \begin{align}
    &{z}_{t}^0\sim\mathcal{N}(\sqrt{\Bar{\alpha}}_{t}{z}^0_0, (1-\Bar{\alpha}_{t})\mathbf{I})
    \enspace, \label{eq:naive_sampling_a}\\
    &\hat{z}_{t}^0\gets {z}_{t}^0
    \enspace, \label{eq:naive_sampling_b} \\
    &\hat{\mathbf{z}}_{t-1}\sim\mathcal{N}(\mu_\theta(\hat{\mathbf{z}}_t, y), \sigma_t^2\mathbf{I})\enspace.
    \label{eq:naive_sampling_c}
    \end{align}
\end{subequations}
\end{small}
Specifically, in each reverse step from $\hat{\mathbf{z}}_{t}$ to $\hat{\mathbf{z}}_{t-1}$, as shown in \cref{eq:naive_sampling_a}, we first compute the noisy latent ${z}_{t}^0$ by adding Gaussian noise to the given image latent ${z}_{0}^0$ over $t$ steps. 
Then, we replace the first latent $\hat{z}_{t}^0$ of $\hat{\mathbf{z}}_{t}$ with ${z}_{t}^0$ in \cref{eq:naive_sampling_b} to incorporate the provided image into the generation process.
Finally, in \cref{eq:naive_sampling_c}, we pass $\hat{\mathbf{z}}_{t}$ through the denoising network to generate $\hat{\mathbf{z}}_{t-1}$, where the text $y$ is integrated by classifier-free guidance (\cref{eq:classifier}).
After $T$ iterations, the final clean latent $\hat{\mathbf{z}}_0$ at diffusion step 0 can be mapped back into the image space $\mathcal{X}$ using the decoder $\mathcal{D}$.

Using this replacing-based baseline, we might expect that the temporal attention layers in $\epsilon_\theta$ can utilize the context provided by the first frame latent $\hat{z}_t^0$ to generate the subsequent frame latents in a manner that harmonizes with $\hat{z}_t^0$.
However, as shown in Fig.~\ref{fig:motivation}, row~2, this replacing-based approach fails to produce a video that is temporally consistent with the first image. The generated frames are consistent with each other, but not with the provided first frame. 

To analyze possible reasons for failure, we apply this baseline to a simpler video infilling task, where every other frame is provided and the model needs to predict the interspersed frames.
In this case, the baseline replaces the generated frame latents at positions corresponding to real frames with noisy provided-frame latents in each reverse step.
The resulting video, in \cref{fig:motivation}, row 3, looks like a combination of two independent videos: the generated (even) frames are consistent with each other but not with the provided (odd) frames.
We speculate that this may result from the intrinsic dissimilarity between frame latents derived from the given real images and those sampled from $\epsilon_\theta$.
Thus, the temporal attention values between frame latents sampled in the same way (both from the given images or both from $\epsilon_\theta$) will be higher, while the attention values between frame latents sampled in different ways (one from the given image and the other from $\epsilon_\theta$) will be lower. 
Therefore, the temporal attention layers of $\mathcal{M}$ tend to utilize the information from latents produced by $\epsilon_\theta$ to synthesize new frames at each reverse step, ignoring the provided frames. 
We further simplify the task to single-frame prediction, where the model only needs to predict a single frame when all the other frames in the video are given. 
In this setting, all the frame latents except for the final frame are replaced by noisy provided-frame latents in each reverse step. Thus, temporal attention layers can only use information from the real frames.
In this case, \cref{fig:motivation}, row 4, shows that the baseline can now generate a final frame that is consistent with the previous frames.

\begin{figure}[t]
    \centering
    \includegraphics[width=0.9\linewidth]{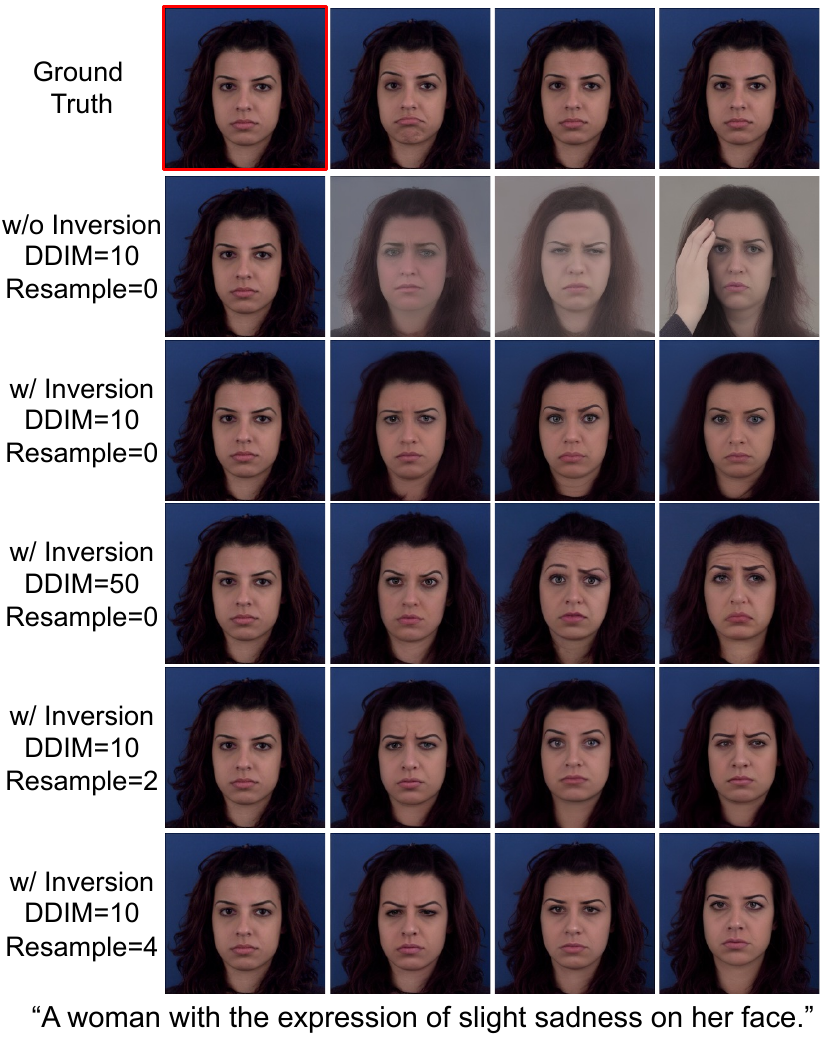}
    \caption{Qualitative ablation study comparing different sampling strategies for our TI2V-Zero on MUG. The first image $\hat{x}^0$ is highlighted with the red box and text $y$ is shown under the block. The 1st, 6th, 11th, and 16th frames of the videos are shown in each column.
    The terms \texttt{Inversion}, \texttt{DDIM}, and \texttt{Resample} denote the application of DDPM inversion, the steps using DDIM sampling, and the iteration number using resampling, respectively.}
    \label{fig:ablation}
    \vspace{-1em}
\end{figure}

\textbf{Repeat-and-Slide Strategy.} Inspired by the observation in \cref{fig:motivation}, to guarantee that the temporal attention layers of $\mathcal{M}$ depend solely on the given image, we make two major changes to the proposed replacing-based baseline: 
(1) instead of using $\mathcal{M}$ to directly synthesize the entire $(K+1)$-frame video, we switch to a frame-by-frame generation approach, i.e., we generate only one new frame latent in each complete DM sampling process; 
(2) for each sampling process generating the new frame latent, we ensure that only one frame latent is produced from $\epsilon_\theta$, while the other $K$ frame latents are derived from the given real image and previously synthesized frames, thereby forcing temporal attention layers to only use the information from these frame latents. 
Specifically, we construct a queue of $K$ frame latents, denoted as $\mathbf{s}_0=\langle s^0_0, s^1_0, \cdots, s^{K-1}_0\rangle$. We also define  $\mathbf{s}_t=\langle s^0_t, s^1_t, \cdots, s^{K-1}_t\rangle$, which is obtained by adding $t$ steps of Gaussian noise to the clean $\mathbf{s}_0$. Similar to our replacing-based baseline in the single-frame prediction task, in each reverse step from $\hat{\mathbf{z}}_t$ to $\hat{\mathbf{z}}_{t-1}$, we replace the first $K$ frame latents in $\hat{\mathbf{z}}_t$ by $\mathbf{s}_t$.
Consequently, the temporal attention layers have to utilize information from $\mathbf{s}_0$ to synthesize the new frame's latent, $\hat{z}^{K}_0$.
Considering that only one starting image latent $z^0$ is provided, we propose a ``repeat-and-slide'' strategy to construct $\mathbf{s}_0$. At the beginning of video generation, we \textit{repeat} $z^0$ for $K$ frames to form $\mathbf{s}_0$, and gradually perform a \textit{sliding} operation within the queue $\mathbf{s}_0$ by dequeuing the first frame latent ${s}^0_0$ and enqueuing the newly generated latent $\hat{z}^K_0$ after each complete DM sampling process. 
Note that though the initial $\mathbf{s}_0$ is created by repeating $z^0$, the noise added to get $\mathbf{s}_t$ is different for each frame's latent in $\mathbf{s}_t$, thus ensuring diversity.
The following expressions define one reverse step in the DM sampling process:
\begin{small}
\begin{subequations}
\label{eq:repeating_and_sliding}
    \begin{align}
        &\mathbf{s}_{t}\sim\mathcal{N}(\sqrt{\Bar{\alpha}}_{t}\mathbf{s}_0, (1-\Bar{\alpha}_{t})\mathbf{I})\enspace, \label{eq:repeating_and_sliding_a}\\
        &\langle\hat{z}_{t}^0, \hat{z}_{t}^1, \cdots, \hat{z}_{t}^{K-1}\rangle\gets \mathbf{s}_{t}\enspace,\label{eq:repeating_and_sliding_b}\\
        &\hat{\mathbf{z}}_{t-1}\sim\mathcal{N}(\mu_\theta(\hat{\mathbf{z}}_t, y), \sigma_t^2\mathbf{I})\enspace.\label{eq:repeating_and_sliding_c}
    \end{align}
\end{subequations}
\end{small}
Specifically, in each reverse denoising step from $\hat{\mathbf{z}}_{t}$ to $\hat{\mathbf{z}}_{t-1}$, we first add $t$ steps of Gaussian noise to the queue $\mathbf{s}_0$ to yield $\mathbf{s}_t$ in \cref{eq:repeating_and_sliding_a}. Subsequently, we replace the previous $K$ frames of $\hat{\mathbf{z}}_{t}$ with $\mathbf{s}_t$ in \cref{eq:repeating_and_sliding_b} and input $\hat{\mathbf{z}}_{t}$ to the denoising network to produce the less noisy latent $\hat{\mathbf{z}}_{t-1}$ (\cref{eq:repeating_and_sliding_c}).

With the repeat-and-slide strategy, model $\mathcal{M}$ is tasked with predicting only one new frame, while the preceding $K$ frames are incorporated into the reverse process to ensure that the temporal attention layers depend solely on information derived from the provided image. 

\textbf{DDPM-based Inversion.} Though the DM sampling process starting with randomly sampled Gaussian noise produces matching semantics, the generated video is often temporally inconsistent (\cref{fig:ablation}, row 2). 
To provide initial noise that can produce more temporally consistent results, we introduce an inversion strategy based on the DDPM \cite{ho2020denoising} forward process when generating the new frame latent.
Specifically, at the beginning of each DM sampling process to synthesize the new frame latent $\hat{z}^{K}_0$, instead of starting with the $\hat{\mathbf{z}}_{T}$ randomly sampled from $\mathcal{N}(\mathbf{0}, \mathbf{I})$, we add $T$ full steps of Gaussian noise to $\mathbf{s}_0$ to obtain $\mathbf{s}_T$ using \cref{eq:single_forward}. 
Note that $\hat{\mathbf{z}}$ has $K+1$ frames, while $\mathbf{s}$ has $K$ frames.
We then use $\mathbf{s}_T$ to initialize the first $K$ frames of $\hat{\mathbf{z}}_{T}$. 
We copy the last frame $s^{K-1}_T$ of $\mathbf{s}_T$ to initialize the final frame $\hat{z}^K_{T}$, as the $(K-1)$th frame is the closest to the $K$th frame.

\textbf{Resampling.} Similar to \cite{lugmayr2022repaint, hoppe2022diffusion}, we further apply a resampling technique, which was initially designed for the image inpainting task, to the video DM to enhance motion coherence. 
Particularly, after performing a one-step denoising operation in the reversed process, we add one-step noise again to revert the latent. This procedure is repeated multiple times for each diffusion step, ensuring harmonization between the predicted and conditioning frame latents (see \cref{alg:repeating_and_sliding} for details).

\begin{table}[t]
\centering
\resizebox{\linewidth}{!}{%
\begin{tabular}{c|c|c|c|c|c}
\hline
Inversion & DDIM & Resample & FVD$\downarrow$     & sFVD$\downarrow$           & tFVD$\downarrow$          \\
\hline
\xmark              & 10     & 0            & 1656.37 & 2074.77\textpm411.74 & 1798.05\textpm235.34 \\
\hline
\cmark              & 10     & 0            & 339.89  & 443.97\textpm139.10  & 405.22\textpm61.58   \\
\cmark              & 50     & 0            & 463.55  & 581.32\textpm234.09  & 535.06\textpm85.27   \\
\hline
\cmark             & 10     & 2            & 207.62  & 299.14\textpm87.24   & 278.73\textpm47.84   \\
\cmark              & 10     & 4            & \textbf{180.09}  & \textbf{267.17\textpm74.72}   & \textbf{252.77\textpm39.02}  \\
\hline
\end{tabular}%
}
\caption{Quantitative ablation study comparing different sampling strategies
for proposed TI2V-Zero on the MUG dataset. \texttt{Inversion}, \texttt{DDIM}, and \texttt{Resample} denote the application of DDPM-based inversion, the steps using DDIM sampling, and the iteration number using resampling, respectively.}
\label{tab:aba}
\end{table}

\begin{table}[t]
\centering
\resizebox{0.9\linewidth}{!}{%
\begin{tabular}{l|c|c}
\hline
Distributions for Comparison            & FVD$\downarrow$    & tFVD$\downarrow$                             \\
\hline
TI2V-Zero-Fake \textit{vs.} ModelScopeT2V & \textbf{366.41} & \textbf{921.31\textpm251.85}  \\
TI2V-Zero-Real \textit{vs.} Real Videos   & {477.19} & {1306.75}\textpm{271.82} \\
\hline
ModelScopeT2V \textit{vs.} Real Videos   & 985.82 & 2264.08\textpm501.28 \\
TI2V-Zero-Fake \textit{vs.} Real Videos   & \textbf{937.11} & \textbf{2177.70\textpm436.71} \\
\hline
\end{tabular}%
}
\caption{Result analysis of TI2V-Zero starting from the real (i.e., TI2V-Zero-Real) or synthesized frames (i.e., TI2V-Zero-Fake) on the UCF101 dataset.}
\label{tab:fake}
\vspace{-1em}
\end{table}

\begin{figure*}[t]
    \centering
    \includegraphics[width=\textwidth]{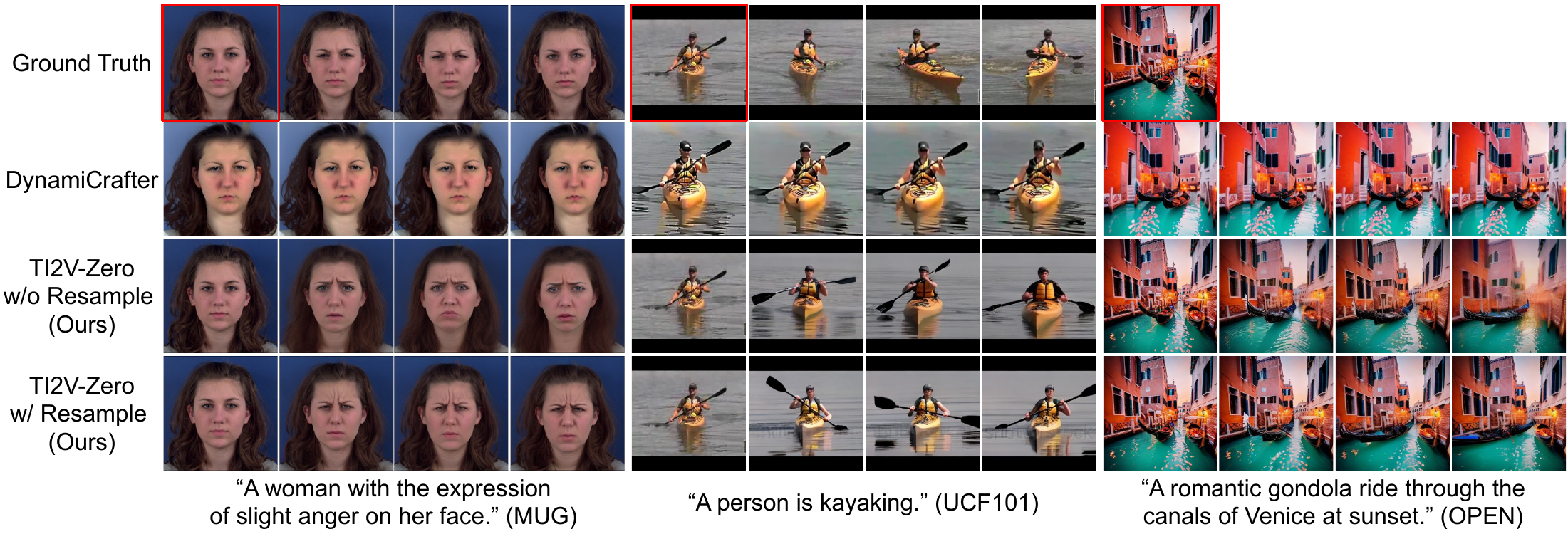}
    \caption{Qualitative comparison among different methods on multiple datasets for TI2V generation. Columns in each block display the 1st, 6th, 11th, and 16th frames of the output videos, respectively. There are 16 frames with a resolution of $256\times256$ for each video. The given image $x^0$ is highlighted with the red box and the text prompt $y$ is shown under each block. 
    }
    \label{fig:sota}
\end{figure*}

\begin{table*}[t]
\centering
\resizebox{0.8\textwidth}{!}{%
\begin{tabular}{l|ccc|cc}
\hline
\multirow{2}{*}{Model}                                             & \multicolumn{3}{c|}{MUG}                                                            & \multicolumn{2}{c}{UCF101}                  \\ \cline{2-6} 
& \multicolumn{1}{c|}{FVD$\downarrow$}     & \multicolumn{1}{c|}{sFVD$\downarrow$}           & tFVD$\downarrow$           & \multicolumn{1}{c|}{FVD$\downarrow$}    & tFVD$\downarrow$           \\ \hline
DynamiCrafter \cite{xing2023dynamicrafter}                                                     & \multicolumn{1}{c|}{1094.72} & \multicolumn{1}{c|}{1359.86\textpm257.73} & 1223.89\textpm105.94 & \multicolumn{1}{c|}{589.59} & 1540.02\textpm{199.59} \\ \hline
TI2V-Zero w/o Resample (Ours) & \multicolumn{1}{c|}{339.89}  & \multicolumn{1}{c|}{443.97\textpm139.10}  & 405.22\textpm61.58   & \multicolumn{1}{c|}{493.19} & 1319.77\textpm283.87 \\
TI2V-Zero w/ Resample (Ours)                                                         & \multicolumn{1}{c|}{\textbf{180.09}}  & \multicolumn{1}{c|}{\textbf{267.17\textpm74.72}}   & \textbf{252.77\textpm39.02}   & \multicolumn{1}{c|}{\textbf{477.19}} & \textbf{1306.75\textpm271.82} \\ \hline
\end{tabular}%
}
\caption{Quantitative comparison among different methods on multiple datasets for TI2V generation.}
\label{tab:sota}
\end{table*}

\section{Experiments}
\subsection{Datasets and Metrics}
We conduct comprehensive experiments on three datasets. More details about datasets, such as selected subjects and text prompts, can be found in our Supplementary Materials.

\textbf{MUG} facial expression dataset \cite{aifanti2010mug} contains 1,009 videos of 52 subjects performing 7 different expressions. We include this dataset to evaluate the performance of models in scenarios with small motion and a simple, unchanged background. 
To simplify the experiments, we randomly select 5 male and 5 female subjects, and 4 expressions.
We use the text prompt templates like ``\texttt{A woman with the expression of slight \{label\} on her face.}'' to change the expression class label to be text input. Since the expressions shown in the videos of MUG are often not obvious, we add ``slight'' in the text input to avoid large motion. 

\textbf{UCF101} action recognition dataset \cite{soomro2012ucf101} contains 13,320 videos from 101 human action classes. We include this dataset to measure performance under complicated motion and complex, changing backgrounds. To simplify the experiments, we select 10 action classes and the first 10 subjects within each class.
We use text prompt templates such as ``\texttt{A person is performing \{label\}}.'' to change the class label to text input. 

In addition to the above two datasets, we create an \textbf{OPEN} dataset to assess the model's performance in open-domain TI2V generation. We first utilize ChatGPT \cite{openaichatgptblog} to generate 10 text prompts. Subsequently, we employ Stable Diffusion 1.5 \cite{rombach2022high} to synthesize 100 images from each text prompt, generating a total of 1,000 starting images and 10 text prompts for evaluating TI2V models. 

\textbf{Data Preprocessing.} We resize all the videos/images to $256\times 256$ resolution. For UCF101, since most of the video frames are not square, we crop the central part of the frames. To obtain ground truth videos for computing metrics, we uniformly sample 16 frames from each video in the datasets to generate the video clips with a fixed length.  

\textbf{Metrics.} 
Following prior work \cite{ho2022imagen, ho2022video, hu2022make}, 
we assess the \textit{visual quality}, \textit{temporal coherence}, and \textit{sample diversity} of generated videos using Fr\'echet Video Distance (\textbf{FVD}) \cite{unterthiner2018towards}.
Similar to Fr\'echet Inception Distance (FID) \cite{heusel2017gans}, which is used for image quality evaluation, FVD utilizes a video classification network I3D \cite{carreira2017quo} pretrained on Kinetics-400 dataset \cite{kay2017kinetics} to extract feature representation of real and synthesized videos. 
Then it calculates the Fr\'echet distance between the distributions of the real and synthesized video features. 
To measure how well a generated video aligns with the text prompt $y$ (\textit{condition accuracy}) and the given image $x_0$ (\textit{subject relevance}), following \cite{ni2023conditional}, we design two variants of FVD, namely text-conditioned FVD (\textbf{tFVD}) and subject-conditioned FVD (\textbf{sFVD}). 
tFVD and sFVD compare the distance between real and synthesized video feature distributions under the same text $y$ or the same subject image $x_0$, respectively. 
We first compute tFVD and sFVD for each condition $y$ and image $x_0$, then report their mean and variance as final results.
In our experiments, we generate 1,000 videos for all the models to estimate the feature distributions. 
We compute both tFVD and sFVD on the MUG dataset, but for UCF101, we only consider tFVD since it doesn't contain videos of different actions for the same subject. 
For the OPEN dataset, we only present qualitative results due to the lack of ground truth videos. Unless otherwise specified, all the generated videos are 16 frames (i.e., $M=15$) with resolution $256\times 256$.

\subsection{Implementation Details}

\hspace{\parindent}\textbf{Model Implementation.} 
We take the ModelScopeT2V 1.4.2 \cite{wang2023modelscope} as basis and implement our modifications. 
For text-conditioned generation, we employ classifier-free guidance with $g=9.0$ in \cref{eq:classifier}. Determined by our preliminary experiments, we choose 10-step DDIM and 4-step resampling as the default setting for MUG and OPEN, and 50-step DDIM and 2-step resampling for UCF101.

\textbf{Implementation of SOTA Model.} We compare our TI2V-Zero with a state-of-the-art (SOTA) model \textit{DynamiCrafter}, a recent open-domain TI2V framework \cite{xing2023dynamicrafter}. DynamiCrafter is based on a large-scale pretrained T2V foundation model VideoCrafter1 \cite{he2022latent}. It introduces a learnable projection network to enable image-conditioned generation and then fine-tunes the entire framework. We implement DynamiCrafter using their provided code with their default settings. 
For a fair comparison, all the generated videos are centrally-cropped and resized to $256\times 256$. 

\subsection{Result Analysis}
\hspace{\parindent}\textbf{Ablation Study.} We conduct ablation study of different sampling strategies on MUG. As shown in \cref{tab:aba} and \cref{fig:ablation}, compared with generating using randomly sampled Gaussian noise, initializing the input noise with DDPM inversion is important for generating temporally continuous videos, improving all of the metrics dramatically.
For MUG, increasing the DDIM sampling steps from 10 to 50 does not enhance the video quality but requires more inference time. 
Thus, we choose 10-step DDIM as the default setting on MUG. 
As shown in \cref{fig:ablation} and \cref{tab:aba}, adding resampling helps preserve identity details (\eg, hairstyle and facial appearance), resulting in lower FVD scores. Increasing resampling steps from 2 to 4 further improves FVD scores. 

\begin{figure}[t]
    \centering
    \includegraphics[width=0.88\linewidth]{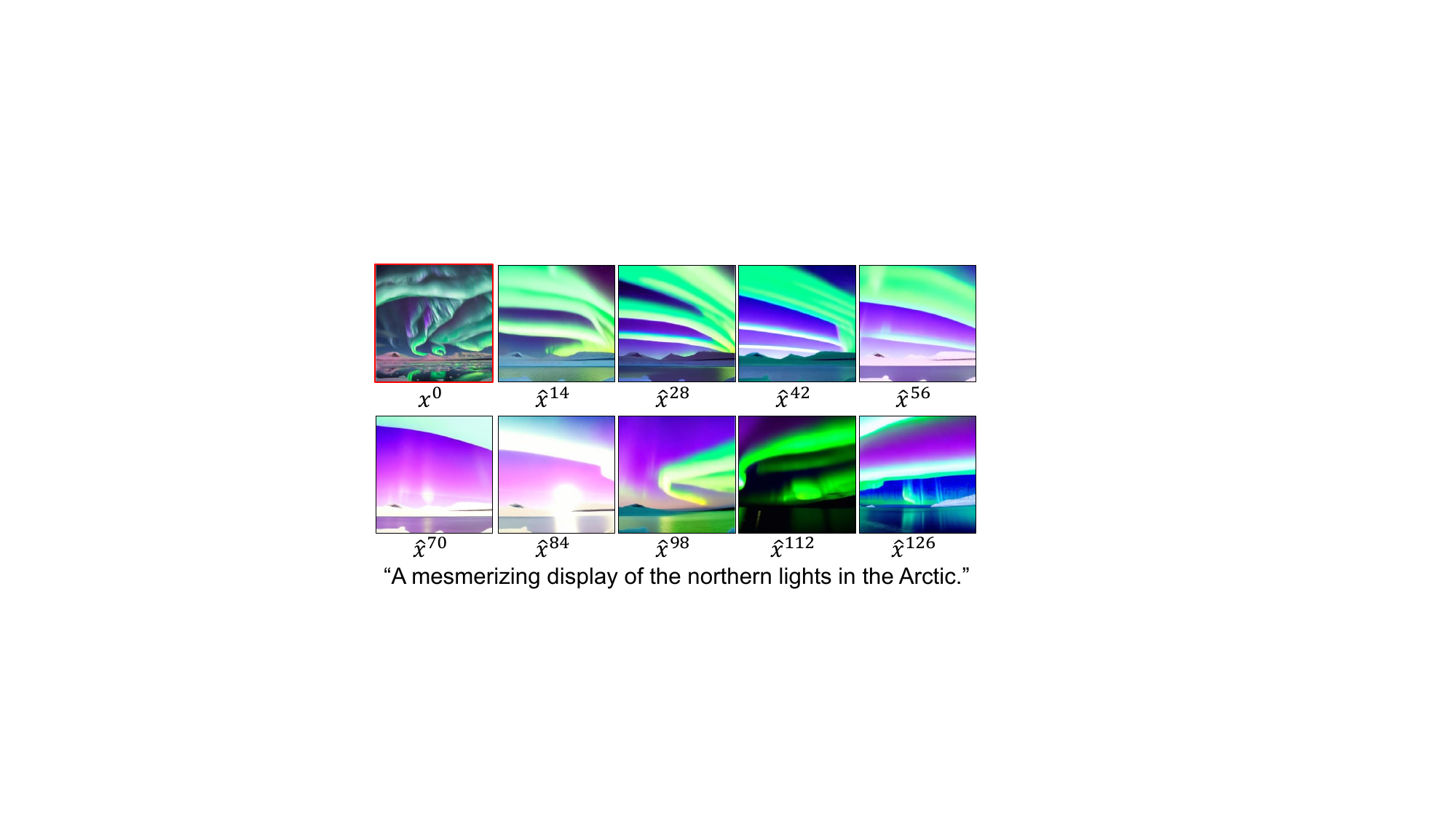}
    \caption{Example of long video generation using our TI2V-Zero on the OPEN dataset. The given image $x^0$ is highlighted with a red box, and the text prompt $y$ is shown under the set of frames. There are a total of 128 video frames ($M=127$), and the synthesized results for every 14 frames are presented.}
    \label{fig:long_video}
    \vspace{-1em}
\end{figure}

\textbf{Effect of Real/Synthesized Starting Frames.} We also explore the effect of video generation starting with real or synthesized frames on UCF101. 
We initially use the first frame of the real videos to generate videos with our TI2V-Zero, termed TI2V-Zero-Real.
Additionally, we utilize the backbone model ModelScopeT2V \cite{wang2023modelscope} to generate synthetic videos using the text inputs of UCF101. We then employ TI2V-Zero to create videos from the first frame of the generated fake videos, denoted as TI2V-Zero-Fake. 
As shown in \cref{tab:fake}, [TI2V-Zero-Fake \textit{vs.} ModelScopeT2V] can achieve better FVD scores than [TI2V-Zero-Real \textit{vs.} Real Videos]. 
The reason may be that frames generated by ModelScopeT2V can be considered as \textit{in-distribution} data since TI2V-Zero is built upon it.
We also compare the output video distribution of TI2V-Zero-Fake and ModelScopeT2V with real videos in \cref{tab:fake}. Though starting from the same synthesized frames, TI2V-Zero-Fake can generate more realistic videos than the backbone model.

\textbf{Comparison with SOTA Model.} We compare our proposed TI2V-Zero with DynamiCrafter \cite{xing2023dynamicrafter} in \cref{tab:sota} and \cref{fig:sota}. From \cref{fig:sota}, one can find that DynamiCrafter struggles to preserve details from the given image, and the motion of its generated videos is also less diverse. Note that DynamiCrafter requires additional fine-tuning to enable TI2V generation. In contrast, without using any fine-tuning or introducing external modules, our proposed TI2V-Zero can precisely start with the given image and output more visually-pleasing results, thus achieving much better FVD scores on both MUG and UCF101 datasets in \cref{tab:sota}. 
The comparison between our TI2V-Zero models with and without using resampling in \cref{fig:sota} and \cref{tab:sota} also demonstrates the effectiveness of using resampling, which can help maintain identity and background details. 

\textbf{Extension to Other Applications.} TI2V-Zero can also be extended to other tasks as long as we can construct $\mathbf{s}_0$ with $K$ images at the beginning. These images can be obtained either from ground truth videos or by applying the repeating operation. Then we can slide $\mathbf{s}_0$ when generating the subsequent frames. We have applied TI2V-Zero in video infilling (see the last row in \cref{fig:motivation}), video prediction (see Supplementary Materials), and long video generation (see \cref{fig:long_video}). As shown in \cref{fig:long_video}, when generating a 128-frame video on the OPEN dataset, our method can preserve the mountain shape in the background, even at the 71st frame (frame $\hat{x}^{70}$). The generated video examples and additional experimental results are in our Supplementary Materials.

\section{Conclusion}
In this paper, we propose a zero-shot text-conditioned image-to-video framework, TI2V-Zero, to generate videos by modulating the sampling process of a pretrained video diffusion model without any optimization or fine-tuning. Comprehensive experiments show that TI2V-Zero can achieve promising performance on multiple datasets.

While showing impressive potential, our proposed TI2V-Zero still has some limitations. First, as TI2V-Zero relies on a pretrained T2V diffusion model, the generation quality of TI2V-Zero is constrained by the capabilities and limitations of the pretrained T2V model. We plan to extend our method to more powerful video diffusion foundation models in the future. Second, our method sometimes generates videos that are blurry or contain flickering artifacts. One possible solution is to apply post-processing methods such as blind video deflickering \cite{lei2023blind} or image/video deblurring \cite{sahu2019blind} to enhance the quality of final output videos or the newly synthesized frame in each generation. Finally, compared with GAN and standard video diffusion models, our approach is considerably slower because it requires running the entire diffusion process for each frame generation. We will investigate some faster sampling methods \cite{kong2021fast,lu2022dpm} to reduce generation time.

{
    \small
    \bibliographystyle{ieeenat_fullname}
    \bibliography{main}
}


\end{document}


\title{Supplementary Materials for\\
TI2V-Zero: Zero-Shot Image Conditioning for Text-to-Video Diffusion Models}  

\author{
Haomiao Ni$^1$\thanks{Work done during an internship at MERL.}
\qquad Bernhard Egger$^2$
\qquad Suhas Lohit$^3$
\qquad Anoop Cherian$^3$
\qquad Ye Wang$^3$\\
Toshiaki Koike-Akino$^3$
\qquad Sharon X. Huang$^1$
\qquad Tim K. Marks$^3$ \\
\small{$^1$The Pennsylvania State University, USA 
\qquad$^2$Friedrich-Alexander-Universit\"{a}t Erlangen-N\"{u}rnberg}, Germany\\
\small{$^3$Mitsubishi Electric Research Laboratories (MERL), USA
}\\
\scriptsize{
$^1${\tt\{hfn5052, suh972\}@psu.edu}
\quad$^2${\tt bernhard.egger@fau.de}
\quad$^3${\tt\{slohit,cherian,yewang,koike,tmarks\}@merl.com}
}\\
\small{\url{https://merl.com/demos/TI2V-Zero}}
}

\maketitle
\thispagestyle{empty}
\appendix

\section{Dataset Details}
We conduct extensive experiments on three diverse datasets, including facial expression dataset MUG, action recognition dataset UCF101, and our self-created dataset OPEN. Here we present comprehensive details about these datasets.

For the MUG dataset, we randomly select 5 male and 5 female subjects from the available 52 individuals, and 4 expressions from the provided 7 expression classes. Detailed information about selected subjects and corresponding expression labels are presented in \cref{tab:mug}. To convert expression class labels to text prompts for input, we use the following templates: ``\texttt{A woman with the expression of slight \{label\} on her face.}'' for female subjects, and ``\texttt{A man with the expression of slight \{label\} on his face.}'' for male subjects. Considering that the average original video length on MUG is about 72 frames, we uniformly sample 16 frames from most of the videos to create the real videos. For videos with more than 80 frames, we sample the videos every 5 frames until we obtain 16 frames to form the real videos. 

For the UCF101 dataset, we initially randomly select some action classes from the provided 101 classes. Subsequently, we identify and choose 10 action classes where both ModelScopeT2V and VideoCrafter1 are able to synthesize promising videos. Table \ref{tab:ucf} shows the details of selected action class labels and their corresponding text prompts. For each action class, we simply choose the first 10 subjects. Given that the average original video length on the UCF101 dataset is approximately 200 frames, we sample the videos every 10 frames until 16 frames are obtained to form the real videos. For videos containing less than 160 frames, we uniformly sample 16 frames.

For the OPEN dataset, we first employ ChatGPT 3.5\footnote{\href{https://openai.com/blog/chatgpt}{https://openai.com/blog/chatgpt}} to generate 10 text prompts by inputting the query ``Could you randomly generate 10 text prompts for testing text-to-video models?''. We list these 10 text prompts in \cref{tab:open}.
Then we use Stable Diffusion 1.5 with the model ID \texttt{dreamlike-photoreal-2.0}\footnote{\href{https://huggingface.co/dreamlike-art/dreamlike-photoreal-2.0}{https://huggingface.co/dreamlike-art/dreamlike-photoreal-2.0}} to generate 100 images for each of 10 text prompts, resulting in a total of 1,000 images as starting frames. 

\begin{table}[t]
\centering
\resizebox{0.8\linewidth}{!}{%
\begin{tabular}{l|c}
\hline
Male ID   & 007, 010, 013, 014, 020             \\ \hline
Female ID & 001, 002, 006, 046, 048            \\ \hline
Expression  & Anger, Happiness, Sadness, Surprise \\ \hline
\end{tabular}%
}
\caption{Details of selected subjects and expression classes on the MUG dataset.}
\label{tab:mug}
\end{table}

\begin{table}[t]
\centering
\resizebox{0.96\linewidth}{!}{%
\begin{tabular}{l|l}
\hline
Action Class   & Text Prompt                          \\ \hline
ApplyEyeMakeup & ``A person is applying eye makeup."     \\ \hline
BabyCrawling   & ``A baby is crawling.''                  \\ \hline
BreastStroke   & ``A person is performing breaststroke.'' \\ \hline
Drumming       & ``A person is drumming.''                \\ \hline
HorseRiding    & ``A person is riding horse.''            \\ \hline
Kayaking       & ``A person is kayaking.''                \\ \hline
MilitaryParade & ``Military parade.''                     \\ \hline
PlayingGuitar  & ``A person is playing guitar.''          \\ \hline
Surfing        & ``A person is surfing.''                 \\ \hline
ShavingBeard   & ``A person is shaving beard.''           \\ \hline
\end{tabular}%
}
\caption{Details of selected action class labels and corresponding text prompts on the UCF101 dataset.}
\label{tab:ucf}
\end{table}

\begin{table*}[t]
\centering
\resizebox{0.7\textwidth}{!}{%
\begin{tabular}{c|l}
\hline
1  & \begin{tabular}[c]{@{}l@{}}``A mesmerizing display of the northern lights in the Arctic.''\end{tabular}                       \\ \hline
2  & \begin{tabular}[c]{@{}l@{}}``A bustling street market in Marrakech with colorful textiles and spices.''\end{tabular}          \\ \hline
3  & \begin{tabular}[c]{@{}l@{}}``A futuristic cityscape with holographic advertisements and flying cars.''\end{tabular}           \\ \hline
4  & \begin{tabular}[c]{@{}l@{}}``A romantic gondola ride through the canals of Venice at sunset.''\end{tabular}                   \\ \hline
5  & \begin{tabular}[c]{@{}l@{}}``A group of friends on a road trip, singing along to their favorite songs.''\end{tabular}         \\ \hline
6  & \begin{tabular}[c]{@{}l@{}}``A serene mountain cabin covered in a fresh blanket of snow.''\end{tabular}                       \\ \hline
7  & \begin{tabular}[c]{@{}l@{}}``A thrilling skateboarder performing tricks in a skate park.''\end{tabular}                       \\ \hline
8  & \begin{tabular}[c]{@{}l@{}}``A bustling night market in Bangkok with street food vendors and live music.''\end{tabular}       \\ \hline
9  & \begin{tabular}[c]{@{}l@{}}``A high-speed bullet train racing through a scenic countryside.''\end{tabular}                    \\ \hline
10 & \begin{tabular}[c]{@{}l@{}}``A group of explorers uncovering the mysteries of an ancient temple in the jungle.''\end{tabular} \\ \hline
\end{tabular}%
}
\caption{The 10 text prompts used in the OPEN dataset.}
\label{tab:open}
\end{table*}

\section{Additional Experimental Results}
\hspace{\parindent}\textbf{More Prior Work Comparisons.} In \cref{tab:sota}, we conduct additional experiments to compare our proposed model with the open-domain TI2V model VideoComposer \cite{wang2023videocomposer} on MUG and UCF101 datasets, where our TI2V-Zero achieves superior performance. 

\begin{figure}[t]
    \centering
    \includegraphics[width=0.98\linewidth]{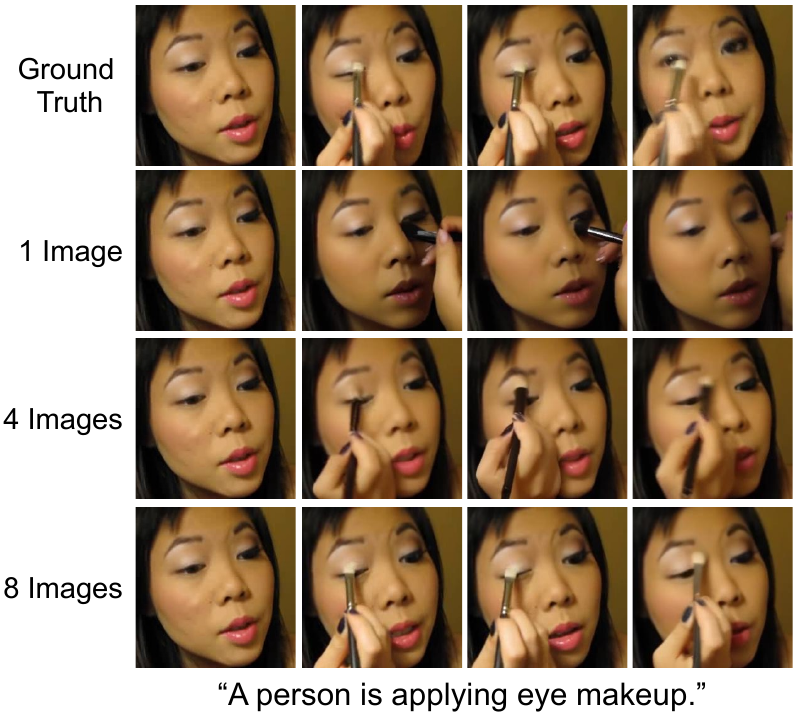}
    \caption{Examples of generated video frames in video prediction task conditioning on different numbers of given images. The 1st, 6th, 11th, and 16th frames of each output video are shown in each column. Each generated video has 16 frames with a resolution of $256\times256$.
    \texttt{1 image}, \texttt{4 images}, \texttt{8 images} indicate the use of the first 1, 4, and 8 real video frames in the ground truth video to predict the next 15, 12, and 8 frames, respectively.}
    \label{fig:prediction}
\end{figure}

\textbf{Extension to Video Prediction Task.} We have presented the results of video infilling and long video generation in the main paper. In \cref{fig:prediction} and our supplementary videos, we show the application of our proposed TI2V-Zero to the video prediction task. Specifically, we conduct experiments using the first 1, 4, and 8 real video frames from the ground truth videos to generate 16-frame videos, i.e., synthesize the subsequent 15, 12, and 8 frames, respectively. As illustrated in \cref{fig:prediction}, when only 1 image is provided, the woman in the generated video applies the powder brush to the eye differently from the real video. With 4 images, the woman in the synthesized video applies the brush to the same eye as in the ground truth video, but it is still hard to maintain the same appearance of the brush as real video. When extending to 8 frames, the model can synthesize a video that is consistent with the given real video.

\begin{table}[t]
\centering
\resizebox{\linewidth}{!}{%
\begin{tabular}{l|ccc|cc}
\hline
\multirow{2}{*}{Model}                                             & \multicolumn{3}{c|}{MUG}                                                            & \multicolumn{2}{c}{UCF101}                  \\ \cline{2-6} 
& \multicolumn{1}{c|}{FVD$\downarrow$}     & \multicolumn{1}{c|}{sFVD$\downarrow$}           & tFVD$\downarrow$           & \multicolumn{1}{c|}{FVD$\downarrow$}    & tFVD$\downarrow$           \\ \hline
VideoComposer \cite{wang2023videocomposer}                                                     & \multicolumn{1}{c|}{1899.08} & \multicolumn{1}{c|}{2294.60\textpm482.45} & 2050.69\textpm116.92 & \multicolumn{1}{c|}{633.32} & 1606.13\textpm{355.87} \\
TI2V-Zero (Ours)                                                         & \multicolumn{1}{c|}{\textbf{180.09}}  & \multicolumn{1}{c|}{\textbf{267.17\textpm74.72}}   & \textbf{252.77\textpm39.02}   & \multicolumn{1}{c|}{\textbf{477.19}} & \textbf{1306.75\textpm271.82} \\ \hline
\end{tabular}%
}
\caption{Quantitative comparison between VideoComposer and TI2V-Zero (w/ Resample) for TI2V generation.}
\label{tab:sota}
\end{table}

\textbf{Inference Time and GPU Usage.} In \cref{tab:inference_time}, we report the average inference time of generating one frame with our proposed TI2V-Zero under different sampling settings, when using a batch size of one on a Quadro RTX 6000 GPU. The GPU usage for each setting is 9,885 MB. With the same GPU, the baseline DynamiCrafter takes about 155 seconds to generate a 16-frame video using their default settings.

\begin{table}[t]
\centering
\resizebox{0.5\linewidth}{!}{%
\begin{tabular}{c|c|c}
\hline
DDIM & Resample & Time (s) \\
\hline
10     & 0          & 5.46                         \\
50     & 0          & 24.86                        \\
\hline
10     & 2          & 14.90                        \\
10     & 4          & 24.70                        \\
\hline
\end{tabular}%
}
\caption{The average inference time for generating one frame using our proposed TI2V-Zero under different sampling settings. The terms \texttt{DDIM} and \texttt{Resample} represent the number of steps of using DDIM sampling and resampling.}
\label{tab:inference_time}  
\end{table}

\section{Discussion with Concurrent Work}
A concurrent work to ours, AnimateZero \cite{yu2023animatezero}, also adopts a similar repeating operation. However, we are different in several aspects. In our framework, when computing temporal attention outputs, the sources of keys are derived either from the given image or previously synthesized images, whereas AnimateZero relies on keys from the given image or noise. Moreover, AnimateZero shares keys and values from spatial self-attention of the first frame across the other frames; this may make it hard to generate large motions and novel scenes, as the content is constrained to the information available in the first frame. In contrast, our framework demonstrates the ability to generate promising videos containing intricate motions with input images of various styles across a wide variety of scenes.

\section{Information about Example Videos}
We include eight MP4 files of example video clips generated by our proposed method in the Supplementary materials. 

\begin{itemize}
    \item \textbf{mug.mp4} includes the video clips generated by the state-of-the-art model DynamiCrafter and our proposed TI2V-Zero for 4 expressions of 4 subjects from the MUG dataset.
    \item \textbf{ucf.mp4} contains the synthesized video clips produced by DynamiCrafter and our TI2V-Zero for action classes from the UCF101 dataset.
    \item \textbf{open.mp4} contains the generated video clips using DynamiCrafter and our TI2V-Zero for 10 text prompts from the OPEN dataset.
    \item \textbf{ablation.mp4} compares the generated video clips under different sampling strategies using our proposed TI2V-Zero on the MUG dataset. The terms \texttt{Inversion}, \texttt{DDIM}, and \texttt{Resample} denote the application of DDPM-based inversion, the steps using DDIM sampling, and the iteration number using resampling, respectively.
    \item \textbf{long\_video.mp4} displays one example video clip showing the application of our proposed TI2V-Zero to generate a 128-frame long video. 
    \item \textbf{prediction.mp4} shows one example video clip illustrating the application of our proposed TI2V-Zero to video prediction task, conditioning on different numbers of given images.
    \item \textbf{motivation.mp4} shows the video clips generated by the replacing-based baseline approach and our proposed TI2V-Zero for different video tasks (corresponding to Fig. 3 in our main paper).
    \item \textbf{intricate.mp4} shows the video clips generated with intricate text and image inputs, including two (16-frame) videos and one long (64-frame) video.  Each first-frame image in the video clips was generated by Stable Diffusion 1.5.
\end{itemize}

{
    \small
    \bibliographystyle{ieeenat_fullname}
    \bibliography{main.bib}
}